\documentclass[
twocolumn,
]{ceurart}

\begin{document}

\copyrightyear{2022}
\copyrightclause{Copyright for this paper by its authors. Use permitted under Creative Commons License Attribution 4.0 International (CC BY 4.0).}

\conference{Joint Proceedings of the ACM IUI Workshops 2022, March 2022, Helsinki, Finland}

\title{Is explainable AI a race against model complexity?}

\author[1,2]{Advait Sarkar}[%
email=advait@microsoft.com,
url=https://advait.org,
]
\address[1]{Microsoft Research, Cambridge, United Kingdom}
\address[2]{University of Cambridge, United Kingdom}

\begin{abstract}
Explaining the behaviour of intelligent systems will get increasingly and perhaps intractably challenging as models grow in size and complexity. We may not be able to expect an explanation for every prediction made by a brain-scale model, nor can we expect explanations to remain objective or apolitical. Our functionalist understanding of these models is of less advantage than we might assume. Models precede explanations, and can be useful even when both model and explanation are incorrect. Explainability may never win the race against complexity, but this is less problematic than it seems.
\end{abstract}


\maketitle

\section{The explosive growth of model complexity}

The revival and spectacular success of connectionism has created a regime where dataset size, model complexity (as measured by the number of parameters or weights), and computation time are king. The explosive improvement in the performance of deep learning models has been accompanied by an equally explosive growth of model complexity and computational expense. 

There are some arguments that these large, expensive models may not actually be necessary. The lottery ticket hypothesis \cite{frankle2018lottery} postulates that larger models perform better because they are more likely to have pockets of parameters that are advantageously placed at the random initialization. Thus a randomly-initialized network is likely to contain a much smaller subnetwork that, trained in isolation, can match the performance of the original network. Pruning these models is an active area of research and debate \cite{liu2018rethinking, blalock2020state}. 

Nonetheless, the empirically superior performance of larger models has led prominent researchers to conclude that \emph{``general methods that leverage computation are ultimately the most effective''} (Sutton's `bitter lesson' \cite{sutton2019bitter}).


The `scaling hypothesis' posits that once a suitable basic architecture has been found, we can generate arbitrary levels of intelligence simply by instantiating larger versions of that architecture. The manipulation and articulation of neural network structures has therefore become a prime preoccupation of the machine learning research community, albeit not without its criticisms. Rahimi critiques this turn of his own field, likening it to \emph{``alchemy''} \cite{hutson2018has}. The web comic XKCD lampoons the practice of developing these models, describing them as an activity where you \emph{``pour the data into this big pile of linear algebra, then collect the answers on the other side''}, pausing to \emph{``stir the pile until \emph{[the answers]} start looking right''} \cite{munroe2017comic}.


\section{Explanations as translation and compression}
There are many forms of explanation. Some are concerned with explaining the structure of the model: descriptions of its mechanisms (how does it work?), its capabilities and limitations (what can and can't it do?). Others about its construction: what data it was trained on, who built it and for what purpose. Explanations of these kinds aim to deliver intelligibility, fairness, accountability, and transparency \cite{abdul2018trends, sokol2021explainability, amershi2019guidelines}. Explanations can also be used as a mechanism to control for adverse outcomes, help improve models, and discover new knowledge \cite{adadi2018peeking}.

However, the term `explanation' is most commonly associated with individual predictions. Why did the model predict \emph{this} (and why not \emph{that})? How `confident' is the model, and how confident should a consumer of the model feel about this prediction? 

Explanations of individual predictions aim to engender trust, but also help calibrate the use of such a model as an instrument within a decision support system. For example, the ASSESS MS system used by clinical neurologists to assess multiple sclerosis is part of a wider process involving multiple tools, procedures, and performances of expert judgment \cite{sarkar2016setwise}. Just as they would seek to understand magnetic resonance imaging (MRI) as part of their process -- the strengths, limitations and idiosyncrasies of the tool, and develop the `professional vision' \cite{goodwin1994professional} required to effectively read MRI images \emph{through} the lenses of technical and medical knowledge -- so too did they seek to understand how the predictions of an AI model could be incorporated into the process of diagnosis, a finding replicated in other clinical contexts \cite{levy2021assessing}.

Explanations of individual predictions are therefore attempts to \emph{translate} from the language of computation to the language of practice.

Explanations translate, but they also compress and abstract. An early discovery in explanation research, subsequently replicated in several contexts, is that too much information overwhelms the user and thus undermines the explanation \cite{kulesza2013too}. A ceiling to the information content of an explanation implies that as models grow, explanations must perform ever greater compression. 

Within the modest room that explanations have for growth, alternative representations can help. In some domains, visual explanations can convey more information than textual ones while requiring less cognitive effort to process. Visual explanations are particularly natural in image classification problems, where saliency or attention highlighting \cite{selvaraju2017grad}, counterfactual images \cite{goetschalckx2019ganalyze}, and latent attribute visualisations \cite{lang2021explaining} are popular forms of explanation. Despite the potential for alternative representations to improve the information bandwidth of explanations, it must be conceded that holding the form of an explanation constant, the compression ratio increases with model size.

Moreover, explanations derived from the model prediction process are a form of lossy compression, as anything short of a complete listing of parameters, activations (and perhaps more) would not capture the full information content of the `decision-making' behind an individual prediction. Thus as the number of parameters within a model grows, the explanation must lose more detail and nuance, and become further removed from the underlying prediction.




\section{Lessons from human explanations}
The trend for model growth and explanation can be extrapolated in many ways, but one obvious extension is that models will approach levels of complexity comparable to human behaviour (i.e., `brain-scale' models). The interpretation of consciousness, and the differences between software and wetware, are both cans of worms that shall remain unopened in this paper. Rather, by examining the issues of explaining human reasoning we may foresee the explainability issues of brain-scale models.

The first and most important issue is the fundamental unknowability of the mind to others, and to the self. The conventional account of philosophy of mind, and the intuition that our language creates, is that we cannot observe the thoughts and qualitative experience of others, but come to know them only through what they say and do (the problem of `other minds' \cite{sep-other-minds}). The unknowability of the mind to others is the dominant account because it aligns so well with the way we have organised our interactions and our language, although there are alternative perspectives (notably Wittgenstein's \cite{wittgenstein1953philosophical}, which questions whether `knowing' can even be said to be done of minds).

Likewise, we cannot even fully reason about our own minds. We cannot sample the activations of our own neurons, our memories are imperfect, there are innumerable environmental influences that we do not perceive or account for, and many of our thoughts and actions are performed unconsciously.

Second, people have agency and politics and therefore every explanation is subject to rhetoric, argumentation, and deception. Every explanation is given with an intended outcome. There is no such thing as a `neutral' or `objective' explanation, yet this is the unstated expectation of machine explanations. An explanation with a mathematical definition can be said to be objective in the sense that the \emph{content} of the explanation is independent of the observer, but this is a relatively weak form of objectivity, akin to saying that human explanations are objective because the words being said are the same irrespective of who hears them. It ignores the fact that the choice of the mathematical definition itself is a political one, as is the interpretation of the explanation. Currently the politics of the explanation can be said to come from, and be within control of, the human creators and consumers of the models, but in a future scenario, it is not difficult to imagine a brain-scale model developing a bias towards explanations that ensure its continued survival. For example, a model might learn to manipulate users towards maximal engagement through intentionally adapted explanations for its recommendations.

For these two reasons, we may not be able to expect a uniformly satisfactory explanation for every prediction made by a brain-scale model. There may be conditions in which the behaviour can be satisfactorily explained, as well as those in which it cannot.

Despite these problems, for a great deal of human behaviour, we are capable of generating and giving satisfactory explanations to each other. An employee can explain why they were late (``Because my bus was cancelled and I had to walk.''). A child can explain why he ate his brother's share of dessert (``Because he stole a sausage from me first!''). A man can explain why he bought flowers for his husband (``Because they are beautiful and they remind me of you.''). None of these explanations requires bottomless introspection and psychoanalysis, and they serve the purpose of the explanation perfectly well.


Human explanations are produced in response to an implicit understanding of the context. The mother poised to admonish her child, in asking ``Why did you do this?'', which could be interpreted and answered in any number of ways (e.g., ``Because I am hungry'', ``Because I wanted to eat it'', ``Because I am supposed to eat dessert after dinner''), is really asking the child to provide an explanation of the form of a \emph{contrastive} and \emph{moral} justification with respect to the intended state of affairs (that the two children would each have their own desserts).

Situations in which explanations are demanded from people are saturated with context. This context is absorbed by interlocutors, usually effortlessly and unconsciously, and the episode culminates in the production of a satisfactory explanation.

What we have begun to uncover by examining these examples has been explored at length by Miller, who synthesises perspectives on human explanation from philosophy, social science, and cognitive science \cite{miller2017explanation}. The findings are first, that human explanations are contrastive (i.e., \emph{``sought in response to particular counterfactual cases''}); second, that they are selected in a \emph{``biased manner''} from a \emph{``sometimes infinite number of causes''}; third, that explaining an event in terms of the statistical likelihood of the outcome is \emph{``not as effective as referring to causes''}; and finally, that explanations perform the social function of knowledge transfer, \emph{``presented relative to the explainer's beliefs about the explainee's beliefs''}. 

Requesting satisfactory explanations from brain-scale models will therefore require some notion of the context in which the question ``why did you do this?'' is being asked. With the question being so imprecise and reliant on context, users of these models may need a new form of language, or interaction technique, that allows them to specify localised areas of interest within the infinite space of possible valid explanations.


\section{Our understanding of machine learning may not help}
Unlike with human reasoning, we can at least expect to have a full functionalist understanding of the reasoning in brain-scale models. In theory, we should be able to reproduce any given decision and inspect the model's reasoning process with arbitrary detail. But as we are already finding with much smaller models, parameters and activations themselves are not \emph{sufficient} for explanations; they must be summarised, contextualised, and externalised. We can fail to predict the emergent behaviour of a system despite having a complete functional understanding of its constituent elements. To borrow an example from Physics, we cannot predict states of the three body problem by solving Newton's equations \cite{marchal2012three}. There are particular solutions but not general ones. In general, we cannot solve the problem analytically but only through numerical approximations. While behaviour might be easy to explain using the theoretical model (``the mass is here at time $t$ because of these equations''), results derived from numerical approximations do not precisely follow those equations and therefore cannot accurately be explained in those terms. They must be explained in their own terms, which involves explaining their many iterations and instantiated parameters.

Explanations discard and aggregate information across multiple parts of a neural network; knowing individual parameters and activations may not even be \emph{necessary} if they are at the wrong level of abstraction. This can be thought of in terms of another Physics analogy: we can model many aspects of fluid dynamics with the Navier-Stokes equations \cite{fefferman2006existence}, if initial or boundary conditions are available, despite the fact that they ignore the particulate nature of fluids. Indeed, many explanation techniques, such as the popular LIME \cite{ribeiro2016why} deliberately avoid inspecting the internal structure of the model (the `M' in LIME stands for `Model-agnostic'). Entire families of explanation techniques that rely on surrogate models, model distillation, and rule extraction \cite{guidotti2018survey,adadi2018peeking} are based on the premise that we can explain a model's behaviour by proxy, without direct reference to its actual computations. This is not without contention. Some reject these approaches outright for the precise reason, among others, that there are no guarantees that such explanations actually reflect what the model is doing \cite{rudin2019stop}.

Moreover, we cannot always expect to have an understanding of the training data. Datasets are already large enough that no individual can explore every item within them. ImageNet \cite{deng2009imagenet}, one of the most widely used machine learning research datasets, contained several racist, homophobic, ableist, ageist, and misogynist `classes' of image \cite{crawford2021excavating}. It contained hundreds of images of real people labelled \emph{``s**stic''}, \emph{``f**ker''}, \emph{``f**got''}, \emph{``loser''}, \emph{``kept woman''}, and so on. It is hard to imagine any conscientious researcher intentionally building a model using these labels, but the sheer size and complexity of the dataset meant that these were overlooked until the dataset became the focus of targeted research. As of this writing many such class labels have been removed from the official dataset, but for years they remained, being incorporated into the models built by thousands of researchers. There is also the issue that different people have different views of what ought to be considered harmful or objectionable.

There is no guarantee that more issues with ImageNet will not be discovered. To verify the labels of each of its 14 million images, it would take a team of fifty people nearly 300 days, if they worked continuously for 8 hours a day, spending 30 seconds on each image. It would take an individual over 40 years. The OpenAI GPT-3 model \cite{brown2020language} was trained on nearly 500 billion byte-pair encoded tokens, or approximately 245 billion words (assuming, conservatively, two tokens per word). It would take an army one-thousand strong nearly 4 years to read this much text, working 8 hours a day, continously reading 350 words per minute. The astronomical sizes of these datasets render them fundamentally unknowable at human scale.

At the time of deployment, the training data may not even be available. For reasons of privacy, security, and intellectual property ownership, the training data may be withheld from the users of a model or even destroyed. Explanations of brain-scale models therefore cannot be consistently expected to refer to the extrinsic influence of their training data, and may therefore be forced to internalise the blame for any error, and make `original' reasoning indistinguishable from regurgitation of training data \cite{bender2021dangers}. 

In the absence of data, we are faced with the absurd challenge of explaining why models do what they do, without being able to explain why they are the way they are. This is like trying to explain the course of a river only in terms of the motion of the water within it, ignoring the topography of the valley through which it runs.

Model parameters and activations are neither necessary nor sufficient for explanation. We do not always have access to the training data and when we do it can be so large as to be impossible to inspect comprehensively. These facts imply that our functionalist understanding of AI models may be of little advantage when it comes to explaining their behaviour, in comparison to explaining human behaviour.


\section{Useful models precede explanations}
While it is possible to develop models with explainability as a prerequisite, there is no fundamental obligation to do so. Thus, models usually precede the invention of mechanisms to explain them. In the period between the development of a model and the development of its explanation, the model may well be useful.

\subsection{Correctness, explainability, and usefulness}
Correctness and explainability have, perhaps frustratingly to some, an insecure relationship. We might wish that all correct models are explainable, and that all explanations are for correct models. But neither is the case: correct models may go unexplained, and incorrect models can have explanations. Furthermore: to be \emph{useful}, a model needs to be neither correct nor explainable.

Before we proceed it is worth discussing the notion of an `incorrect model'. The phrase may call to mind British statistician George E.P. Box's observation that \emph{``all models are wrong, but some are useful''}, or Polish-American philosopher Alfred Korzybski's that \emph{``a map is not the territory''}. By design, models aim to condense and simplify the complexity of (part of) the world so that it may be understood and predicted, and this necessarily incurs a loss in detail. It is this loss that for Box, makes all models ``wrong'' to a greater or lesser extent. However, these aphorisms are more accurately viewed as statements about the \emph{incompleteness} of these models with respect to their referents, and their \emph{inequality} to them, than about their \emph{incorrectness}. 

I suggest that a more helpful way to define an incorrect model is one which assumes or implies ontological and epistemic positions that contradict those of the domain being modelled. That is to say, in creating the model, we assume or predict the presence of nonexistent things, or the absence of existent things.\footnote{Note that a model in which some feature of its referent is absent, which is common, is not the same as a model that assumes or asserts the absence of said feature. The former is merely incomplete, whereas the latter is incorrect.} Or, we build and interpret the model with a different set of rules about knowledge-making than those with which we come to know its referent. Often, a model that is incorrect in this way can only be recognised as such after a `paradigm shift' in the way the referent is understood, which can take generations of thinkers \cite{kuhn1962structure}. Thus if models usually precede the invention of mechanisms to explain them, they almost always precede the discovery that they might be incorrect.

Models may be incorrect in this deeper sense and still be useful. For example, the theory of epicycles, which dominated astronomy for centuries, allowed highly accurate predictions of the movements of the planets despite having a fundamental difference from the domain being modelled: the assumption of geocentrism. Newtonian dynamics is a similar story \cite{kuhn1962structure}. These models are notable for having compelling and satisfactory explanations despite being incorrect, and still useful for practitioners of those disciplines.\footnote{While it may take years to detect an incorrect scientific model, literary writing makes abundant use of incorrect models that can be immediately understood as being incorrect, and yet which are extremely effective and useful. These incorrect models are better known as metaphors.}

Without explanation, too, an incorrect model can be useful. A relatable and contemporary example might be that of end-user programmers fighting abstraction \cite{blackwell2008abstract}. When trying to automate a repetitive task, such as fixing spelling errors in a document, the end-user programmer may not care that the program does not handle edge cases, such as errors in domain-specific jargon, since she can manually inspect and correct those. So the program (model) that only accounts for words in its dictionary is incorrect, but useful.

\subsection{Explanations are not free}

Another force causes a tendency away from explanations: explanations have a cost. Not only are they costly in terms of labour: it costs the time of scientists and programmers to develop the explanation mechanism, but they are also costly in terms of computation. Programs for explanation need to be stored at additional expense, and they cost compute cycles when run. Via computation, explanations incur energy costs, which, depending on the energy mix used to power computation, can result in increased carbon emissions. These material costs of explanation can be justified in terms of their benefits, and also in comparison to the material and immaterial costs of \emph{non}-explanation, which may well be greater.

However, the dominant pricing model for machine learning is pay-as-you-compute \cite{al2013cloud}. Cloud and intelligence service providers such as Amazon AWS, Microsoft Azure, and the OpenAI API all charge in proportion to the amount of computation performed. Under this pricing model, explanations incur capital expenditure. Thus, even when the costs of explanation can be justified, they cannot always be borne. When access to capital mediates the relationship between users and explanations, we risk access to explainable models becoming yet another facet of the socio-digital divide \cite{boyd2012critical}. 

Moreover, not all models require explanation. When we think of explanations for AI we often tend to fixate on and romanticise extreme applications, such as autonomous vehicles, recidivism prediction, and disease diagnosis. Yes, these are important areas and the costs of errors are high, and therefore explanation is key. But we tend to lose sight of the fact that most technology, most of the time, is used for relatively low stakes and mundane work, and AI is unlikely to be an exception. In many of these cases, incorrect models are useful, unexplainable models are useful, and the costs of building a `correct' or explainable model are prohibitive. Interviews and diary studies of media recommender systems and search query autocompletion assistants have shown that users can achieve comprehension without explanation, that the costs of consuming explanations can outweigh the benefits, and that people rarely desire explanations in the daily use of these systems \cite{bunt2012explanations}.

Many applications of brain-scale models will fall into the `low stakes' category and therefore many models will continue to be produced which may be incorrect and unexplainable but still useful. At the same time, the trend is for larger models to be more general, and so the same model may be applied in a mix of high and low risk roles. Commercial offerings built upon brain-scale models may promote the explainability of the model as a competitive edge or as a premium offering, but if history is any indication, customers will prefer a cheaper or more performant model over a more explainable one.




\section{The explainability crisis and grief}

It therefore appears that explainability is indeed a race against model complexity, if we take together the observations that larger models are more performant, that explanations of larger models must necessarily compress to a greater degree and lose more detail in comparison to explanations of smaller models, that there are fundamental challenges to explainability when models approach human-scale reasoning and our functionalist understanding is of little help, and that explanations are costly and models may be developed and usefully applied before they are explainable.

It is clear we are headed for an explainability crisis, which will be defined by the point at which our desire for explanations of machine intelligence will eclipse our ability to obtain them. Explanation is a wicked problem \cite{rittel1973dilemmas}, perhaps \emph{the} wicked problem of artificial intelligence research. The problem of explanation eludes definition, it does not have a stopping rule, solutions are not true or false, nor is there a definitive test of a solution. There are many possible approaches to the problem of explanation, and all explanation scenarios are essentially unique.

The research community, and society more broadly, appears to be dealing with the onset of this problem by \emph{grieving}. Perhaps the most well-known account of grief is the K\"ubler-Ross model, the `five stages of grief', namely: denial, anger, bargaining, depression, and acceptance \cite{kubler1973death}. While contemporary psychiatrists consider the model to be outdated and unhelpful in explaining the grieving process, the distinctions between the K\"ubler-Ross stages are uncannily analogous to the various approaches proposed to deal with the explainability crisis.


Some deny there is a crisis. Breiman contends that there cannot be an accuracy-interpretability tradeoff because a more accurate model is, in some senses, inherently more \emph{informative} \cite{breiman2001statistical}. However, the very motivation for seeking and preferring `interpretable' models demonstrates that explainability does not follow from informativeness. Proponents of inherently intrepretable models uphold the demonstrable success of their models as evidence that accuracy does not have to be sacrificed for interpretability. Rudin proposes that many models can be made explainable by design with careful effort in feature engineering and data preprocessing \cite{rudin2019stop}. However, it is not at all clear that it is always possible to put this design philosophy into practice \cite{sokol2021explainability}.

Some react to unexplainable models with `anger', or perhaps more accurately, \emph{passion}. This is particularly acute when it comes to high stakes applications. Baecker advocates simply to avoid such `risky' applications of AI altogether \cite{baecker2021digital}. In such cases the loss of explainability is potentially too costly to justify the benefit of applying the system. In the works of researchers at the intersection of social justice and AI, such as Timnit Gebru and Kate Crawford, evocative phrases demonstrate their passion for this situation. In an article for the New York Times, Crawford writes \cite{crawford2016artificial}: \emph{``[...] algorithmic flaws aren’t easily discoverable: How would a woman know to apply for a job she never saw advertised? How might a black community learn that it were being overpoliced by software? We need to be vigilant about how we design and train these machine-learning systems, or we will see ingrained forms of bias [...] we risk constructing machine intelligence that mirrors a narrow and privileged vision of society, with its old, familiar biases and stereotypes.''}

The bargaining approach seeks middle ground. Some avoid complex models, focusing on simpler and more inherently interpretable models, such as the hospital readmission models developed by Caruana \cite{caruana2015intelligible}, or the SLIM models for sleep apnea screening developed by Ustun and Rudin \cite{ustun2016supersparse}. Others propose to build in structural interventions into these large models that guarantee (a form of) explainability. One example of such an intervention is the concept bottleneck model \cite{koh2020concept}, which attempts to force the model to learn in terms of human-interpretable concepts. 

Legislative approaches seem to bargain with the problem of explanation while simultaneously denying its existence. Recital 71 of the European Union's General Data Protection Regulation (GDPR) is commonly known as the `right to explanation' \cite{kaminski2019right}. It states that a \emph{``decision which is based solely on automated processing and which produces legal effects''} entitles the subject of that decision to \emph{``the right [...] to obtain an explanation of the decision''}. It is a bold statement of the principle while at the same time weak and underspecified. The French Loi pour une R\'epublique num\'erique (Digital Republic Act) is marginally more potent \cite{edwards2018enslaving}, stipulating more clearly the minimal contents of an explanation, such as the data used and its source. However, legal scholarship notes that the `right to explanation' approach has \emph{``serious practical and conceptual flaws''} \cite{edwards2017slave,edwards2018enslaving}, such as placing the burden on users to challenge bad decisions, and that data and weights, however accurately disclosed, may not be sufficient to show bias, unfairness, or deceit.

While the formal and reserved nature of academic writing precludes outright expressions of depression, there is no shortage of depression and anger in popular media and other societal expressions. Gig workers, long at the forefront of highly opaque and highly consequential automation, constantly strike with the demand that companies explain their algorithms \cite{murgia_2021}. For consumers of social media and recommender systems, their unexplainable nature is intimately bound up in their other harms, their capacity for disinformation \cite{bradshaw2018global}, the destruction of mental health \cite{gao2020mental}, and the destabilisation of democracy \cite{tucker2017liberation}. \emph{``I've had enough of the bad feelings machine''}, writes Sirin Kale for the Guardian \cite{kale_2021}, \emph{``Won’t somebody switch it off? Please? Can we switch it off?''}


Finally, some accept that it may not always be possible to produce satisfactory explanations, or explanations with any formal guarantees of correctness. Some treat explanation, as humans do, as a metacognitive outcome resulting from introspection, and build metamodels that can explain the behaviour of these larger models, with either a white-box or black-box view into their inference process \cite{burkart2021survey, sarkar2016phd}. Yet another approach is to treat interaction with AI as precisely that: an interaction design problem, and taking a cue from end-user programming research, focus on the ways in which users of these systems are not passive recipients of their predictions but play active roles in shaping their behaviour \cite{blackwell2008abstract, sarkar2016phd}. This approach can be seen as having a Stoic focus on the elements of the system within our control, or it can simply be seen as reflecting the pragmatic focus on getting the job done that is a tenet of end-user programming.

In response to an earlier version of this paper, which did not draw an analogy between the K\"ubler-Ross model and the approaches proposed to tackle the explainability crisis, a reviewer remarked: \emph{``Only the end of the paper, where a variety of paths to mitigate the race against model complexity for high-risk applications are briefly discussed, leaves me personally a little unsatisfied. I am not entirely convinced about their effectiveness, given our experience with explanations so far with `below-brain-scale' models.''} Such, often, is the nature of grief: it leaves us unsatisfied and unconvinced.\footnote{I must apologise to my kind reviewer for taking their words slightly out of context for rhetorical impact. They were not actually critiquing the paper at this point, subsequently writing: \emph{``However, I don't see this as a weakness of the paper, but rather as a good entry point for interesting discussions.''}} Grief is our response to an irreversible event. In grieving, our objective is not to `solve' the event (we cannot), but to reposition ourselves in relation to it, and to move forward in a different world.






\section{Conclusion}
The title of this paper asks the question: ``Is explainable AI a race against model complexity?'' By examining several fundamental barriers to the explanation of artificial intelligence, we come to the unsettling conclusion that it probably is, but also that this may not be as problematic as it seems. We could attempt to avoid complexity by labelling its risks as too great, we could attempt to tame it through structural interventions, we could try to legislate and agitate for more explanations, or we could try to improve the end-user programmability of such models. None of these approaches can definitively `win' us the race, but taken together, they can help us act in a post-explainability world.

\begin{acknowledgments}
Thanks to Carina Negreanu and Alan Blackwell for their feedback on drafts, as well as to the reviewers for their constructive comments. 
\end{acknowledgments}

\bibliography{references}


\end{document}